\documentclass[10pt,twocolumn,letterpaper]{article}

\usepackage[pagenumbers]{wacv}

\usepackage{graphicx}
\usepackage{amsmath}
\usepackage{amssymb}
\usepackage{booktabs}
\usepackage{siunitx}
\usepackage[pagebackref,breaklinks,colorlinks]{hyperref}
\usepackage[capitalize]{cleveref}
\usepackage{threeparttable}
\usepackage{tikz}
\crefname{section}{Sec.}{Secs.}
\Crefname{section}{Section}{Sections}
\Crefname{table}{Table}{Tables}
\crefname{table}{Tab.}{Tabs.}

\newcommand{\suppmat}{\href{https://arxiv.org/src/2601.22861v1/anc/supp.pdf}{supplementary material}}
\newcommand{\suppvideo}[1]{\href{https://arxiv.org/src/2601.22861v1/anc/#1}{3D animation}}

\def\wacvPaperID{0004}
\def\confName{WACV}
\def\confYear{2026}

\ifx\IsSupplementary\undefined
\title{Under-Canopy Terrain Reconstruction in Dense Forests\\ Using RGB Imaging and Neural 3D Reconstruction}
\else
\title{Under-Canopy Terrain Reconstruction in Dense Forests\\ Using RGB Imaging and Neural 3D Reconstruction \\ Supplementary Materials}
\fi

\author{
  Refael Sheffer$^1$ \quad Chen Pinchover$^1$ \quad Haim Zisman$^2$ \quad Dror Ozeri$^1$ \quad Roee Litman \\
  $^1$Rafael Advanced Defense Systems inc., Israel \qquad $^2$Bar-Ilan University, Israel
}

\begin{document}

\twocolumn[{%
\renewcommand\twocolumn[1][]{#1}%
\vspace{-1ex}
\maketitle
\vspace{-2em}
  \newcommand{\teasew}{0.18\textwidth}
  \centering
  \addtolength{\tabcolsep}{-0.5pt}    
  \begin{tabular}{ccccc} 
    \includegraphics[width=\teasew]{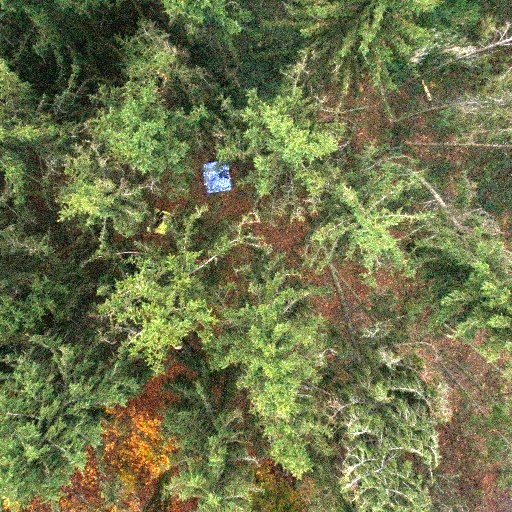} &
    \includegraphics[width=\teasew]{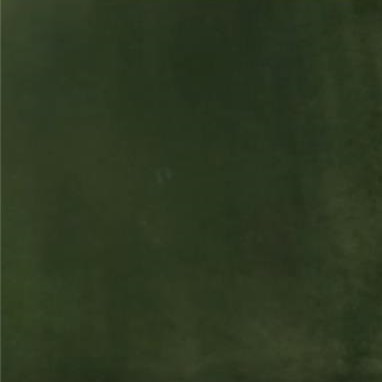} &
    \includegraphics[width=\teasew]{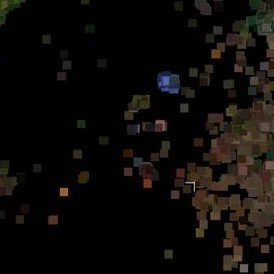} &
    \includegraphics[width=\teasew]{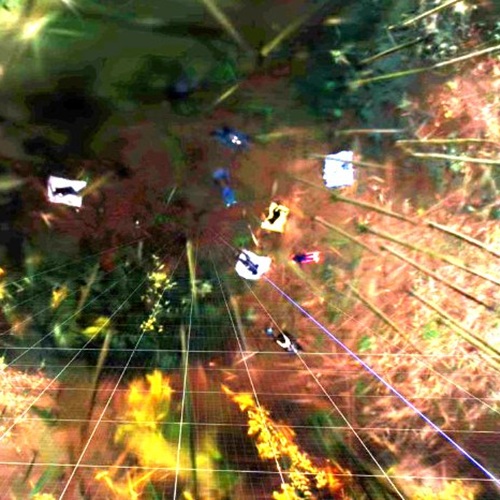} &
    \includegraphics[width=\teasew]{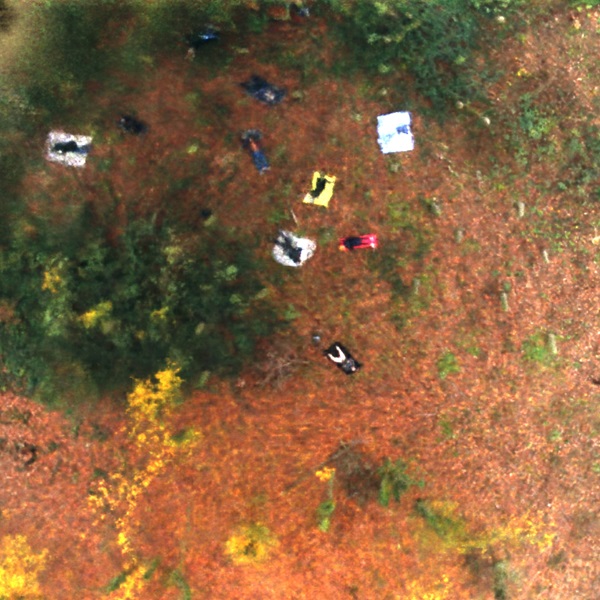} \\
    RGB image \cite{schedl2021airborne} &
    AOS\cite{schedl2021airborne} &
    MVS \cite{schonberger2018robust} & 
    3DGS \cite{lichtfeld2025} &
    Our method \\
  \end{tabular}
  \captionof{figure}{
  \textbf{An example of the several under-canopy imaging methods}
  discussed in this paper, on the `F6' scene from the drone imagery dataset~\cite{schedl2021airborne}. 
  Images are approximately aligned for easy comparison.
  See Section~\ref{sec:quality} for more details.
  }
  \label{fig:teaser}    
  \vspace{2ex}
}]

\begin{abstract}
\small
Mapping the terrain and understory hidden beneath dense forest canopies is of great interest for numerous applications such as search and rescue, trail mapping, forest inventory tasks, and more. 
Existing solutions rely on specialized sensors: either heavy, costly airborne LiDAR, or Airborne Optical Sectioning (AOS), which uses thermal synthetic aperture photography and is tailored for person detection.

We introduce a novel approach for the reconstruction of canopy-free, photorealistic ground views using only conventional RGB images.
Our solution is based on Neural Radiance Fields (NeRF), a recent 3D reconstruction method.
Additionally, we include specific image capture considerations, which dictate the needed illumination to successfully expose the scene beneath the canopy.
To better cope with the poorly lit understory, we employ a low light loss.
Finally, we propose two complementary approaches to remove occluding canopy elements by controlling per-ray integration procedure.

To validate the value of our approach, we present two possible downstream tasks.
For the task of search and rescue (SAR), 
we demonstrate that our method enables person detection which achieves promising results compared to thermal AOS (using only RGB images).
Additionally, we show the potential of our approach for forest inventory tasks like tree counting.
These results position our approach as a cost-effective, high-resolution alternative to specialized sensors for SAR, trail mapping, and forest-inventory tasks.
\end{abstract}

\vspace{-3ex}
\section{Introduction}\label{sec:intro}

Accurate perception of the forest floor beneath dense canopies is a long standing challenge that affects multiple applications such as SAR missions, biodiversity monitoring, understory fuel mapping, forest inventory audits, and carbon stock assessment. 
Current solutions rely on either LiDAR or thermal imaging. 
LiDAR penetrates foliage with active pulses to recover bare earth terrain~\cite{chase2011airborne,evans2013uncovering}, while thermal imaging is based on a recent AOS approach of Schedl \etal~\cite{schedl2020search} which fuses LWIR frames via synthetic aperture to reveal people.

While both technologies are valuable, they leave practical gaps that our approach aims to address.
First, they rely on specialized payloads that tend to be heavy and costly.
Second, these sensors typically yield lower spatial resolution compared to off-the-shelf RGB cameras, limiting the ability to resolve small objects,  terrain features, and subtle undergrowth structures. 
Third, both sensors are typically used on very specific tasks, \eg thermal systems are primarily optimized for detecting warm-bodied targets like people, and are typically less effective for detailed vegetation mapping due to lower contrast relative to the ground.
Last, but not least, beyond hardware and resolution limits, both LiDAR and thermal systems produce data that is inherently less intuitive to interpret, compared to visible light images. 

Beyond hardware constraints, the processing pipelines of these sensors introduce additional challenges.
LiDAR requires specific point cloud processing algorithms to classify ground points, while thermal imagery, especially after SAP, emphasizes hot objects, and suppresses detail in cooler surroundings, often rendering terrain features blurry or indistinct.
In contrast, our NeRF based pipeline trained on RGB from off-the-shelf camera 
outputs an image of the understory, which might include terrain, trails, man made objects, and vegetation gaps across the entire scene. 
Moreover, for thermal imagery data, our method allows mapping and recognition of cool objects (like vegetation) in addition to hot ones\footnote{Results on thermal data are included in the \suppmat{}}.
Leveraging recent advances in implicit volumetric rendering, we train a NeRF model on high overlap images captured by lightweight drones. 
By estimating the canopy and surface, and by restricting volumetric integration bounds accordingly, we synthesize ground only views in which every pixel corresponds to terrain, trails, or stem bases.

\vspace{-3ex}
\paragraph{Our contributions} are fourfold.
First and foremost, we present an approach to produce an image of the terrain under a dense canopy, which is centered around differentiable rendering reconstruction, and using data from a commodity camera.
Second, we formalize the lighting considerations under which such method can be used, and suggest an appropriate loss function to better cope with low light conditions.
Third, we propose two complementary methods for synthesizing under canopy images from the NeRF reconstruction, either using semantic filtering or volumetric clipping using a provided digital terrain model (DTM).
Lastly, we propose two possible downstream applications which are based on our scene reconstruction result, namely person detection and tree counting.

\section{Related Work}

We begin by covering three existing approaches of mapping terrain under the canopy, namely LiDAR, thermal SAP, and classic RGB aerial photogrammetry.
Next, we cover the related works that are the basis of our method, namely differentiable volumetric rendering and semantic segmentation.

\vspace{-2ex}
\paragraph{LiDAR based mapping.}
The physical properties of LiDAR sensors make them perfect for penetrating occlusion such as dense vegetation, and the attempts to map such hidden structures go back at least three decades \cite{kraus1998determination,drake2002estimation,axelsson2000dem}.
Successful cases started appearing with the increase of hardware performance\footnote{both sensor quality and compute power} a decade later \cite{chase2011airborne,evans2013uncovering}, and it is still an active area of research \cite{campbell2018quantifying,venier2019modelling,vanvalkenburgh2020lasers,cmielewski2021uav}. 
While LiDAR sensors become more ubiquitous with additional commercial use cases (such as autonomous vehicles and robotics), they still fall short compared to visible light cameras in terms of price, weight, robustness, and power consumption.
At the time of writing, DJI~\cite{djistore} offers only a limited number of LiDAR modules—such as the Zenmuse L1 and L2—designed exclusively for high-end industrial drones like the Matrice 300/350 RTK. In contrast, a wide range of camera only platforms are available across DJI’s consumer and prosumer lines, offering significantly lower cost and comparable durability for aerial imaging tasks.

\vspace{-2ex}
\paragraph{Thermal under canopy imaging.}
The use of thermal imaging for canopy penetration is much more recent, compared to the aforementioned LiDAR.
A recent line of works uses synthetic aperture photography (SAP) on LWIR thermal images to reveal occluded targets~\cite{schedl2021airborne}. 
This approach essentially refocuses the rays of the camera onto the ground plane, allowing detection of objects that have high contrast compared to the background, \eg people under the canopy.
Thermal SAP is less suitable, however, to map low contrast objects like trails and trees, which typically have temperatures similar to the scene around them.
The reason for this is that SAP essentially doesn’t remove the foliage, it simply `buries' them under the dominant signal of the hotter targets.

While thermal sensors are much cheaper and lighter than LiDAR, they remain more expensive and heavier than standard RGB cameras.
Additionally, thermal sensors typically offer coarser ground sample distance (GSD), limiting fine scale terrain analysis.

\vspace{-2ex}
\paragraph{Aerial photogrammetry and MVS‐based reconstruction.}
Prior to the advent of differentiable rendering methods for 3D reconstruction (see below), classical photogrammetry relied on structure from motion (SfM) and multi view stereo (MVS) to build a dense point cloud. 
One such recent work was presented by Thiel \etal~\cite{thiel2020uas}, where a canopy free orthomosaic is generated from an MVS based point cloud, even though LiDAR terrain mapping is used to filter out the canopy. 
Importantly, such approaches work mostly in leaf off conditions, and struggle in leaf on.

Another approach, more similar to AOS~\cite{schedl2021airborne}, was presented by Ryu \etal~\cite{ryu2023enhanced}, which attempts to recreate the results from LWIR images with RGB with the introduction of HSV based anomaly detection. 
This method works well for partially exposed targets whose color is significantly different from the background and the foliage, but requires the target to have anomalous color. 
Our method, conversely, reconstructs an image of the surface beneath the canopy, and thus can also detect objects based on their shape and texture (as done in `standard` deep learning approaches).

\vspace{-2ex}
\paragraph{Obstruction free photography.}
A seminal work by Xue \etal~\cite{Xue2015ObstructionFree} started a line of works (\eg \cite{li2021let,chugunov2024neural,zhu2023occlusion}) that attempts to remove occluding elements in real world scenes using multi view photography.  
We mention this line of works only for completeness, and do not compare thereto, as it is based on the assumption that the occlusion is a planar 2D layer, and thus can be easily separated from the scene by its parallax. 
This assumption is violated in the case of dense vegetation, as it is both 3D in nature, and is not physically separated from the ground underneath.

\vspace{-2ex}
\paragraph{Differentiable volumetric rendering.}
The world of 3D reconstruction was recently disrupted with the introduction of Neural Radiance Fields (NeRF)~\cite{mildenhall2020nerf}, which was one of the first methods to successfully reconstruct 3D scenes by optimizing a continuous volumetric representation of appearance, without relying on explicit geometry.
NeRF~\cite{mildenhall2020nerf} models the scene using a multi layer perceptron (MLP) network that takes a (continuous) 5D input, which consists 3D position $(x,y,z)$ and 2D viewing direction $(\phi,\theta)$, and outputs color $(r,g,b)$ and density $\sigma$. 
These outputs are integrated along camera rays using volumetric rendering, as described in Eq.~\eqref{eq:nerf} in Section~\ref{sec:nerf}.

NeRF was initially used for novel view synthesis, but many other extensions have arisen since, and some are particularly relevant to our problem setting. 
Recent works like \cite{chugunov2024neural,zhu2023occlusion} marry NeRF with obstruction free approaches~\cite{Xue2015ObstructionFree}.
However, these methods still rely on parallax-based separation of occlusions, which is less effective in dense, volumetric environments like forest canopies.

Beyond NeRF, a notable line of work is 3D Gaussian Splatting (3DGS)~\cite{kerbl20233d}, which replaces the implicit MLP representation with an explicit set of Gaussians distributed in 3D space. 
Each Gaussian encodes position, covariance (first- and second-order moments), and direction-dependent color and opacity, similar to NeRF outputs. 
The optimization process iteratively refines these parameters, adding or removing Gaussians based on rendering quality. 
Compared to NeRF, 3DGS offers significantly faster rendering and provides an explicit 3D structure that is arguably more interpretable. 
However, most 3DGS implementations employ a rasterization-based rendering pipeline rather than NeRF’s ray tracing. 
This distinction is important, as rasterization is still considered less effective for complex, fine-grained geometry and transparency~\cite{akeley2002will,caulfield2018whats}.

\vspace{-2ex}
\paragraph{Semantic image segmentation.}
In this paper, we propose an approach to indicate the occluding canopy, by leveraging semantic segmentation of the input images.
A physical source for such 2D segmentation maps can be obtained using color separation, or even extra spectral data (if available) as done in~\cite{poggi2022xsrnf}, and segment vegetation via an indicator such as NDVI~\cite{pettorelli2013normalized, nasa2000ndvi}
Another option is to use state of the art segmentation model like SAM~\cite{kirillov2023segment}, which requires little to no supervision.
These per-pixel predictions can be aggregated during the 3D model reconstruction process, which will average out any inconsistencies (See Figure 9 in~\cite{rosu2020semi}). 
Several methods like NeRF-SOS~\cite{fan2022nerf} follow this idea, and even segment the 3D scene with no extra supervision based on self supervised models. 
Followup works like \cite{mirzaei2023spin,yin2023or} further remove entire objects from the scene based on such segmentation.

\section{Method}
\begin{figure}[t]
  \centering
  \includegraphics[width=\linewidth]{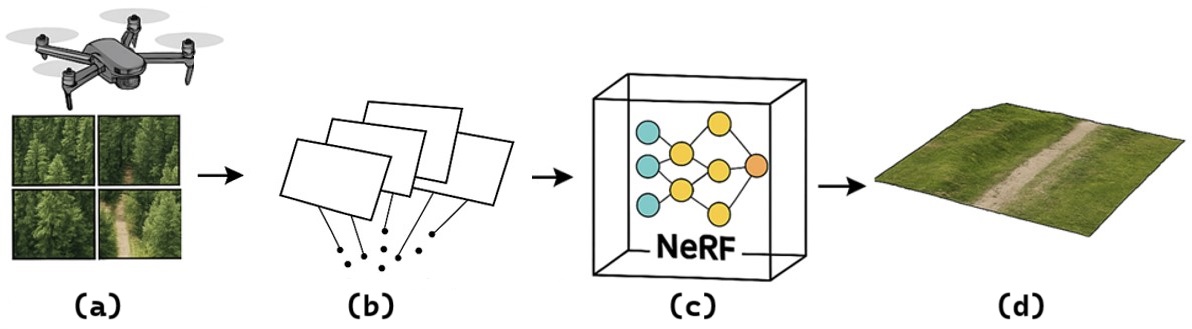}
  \caption{
  \textbf{Proposed method pipeline overview:} (a) High overlap image capture; (b) bundle adjustment;
  (c) NeRF reconstruction; (d) ground only rendering revealing occluded terrain.
  \vspace{-3ex}
  }
  \label{fig:overview}
\end{figure}

Our approach follows a pipeline which comprises four phases: image capture, structure from motion (SfM), NeRF training, and ground only
rendering. Figure \ref{fig:overview} outlines this process.

\subsection{Image Capture Protocol}
We summarize key factors influencing canopy penetration in terms of viewing angle, sampling density, resolution and lighting conditions. 
With the exception of the latter, these guidelines stem from previous works like \cite{schedl2020search, schedl2021autonomous, seits2022role, nathan2024reciprocal} which focus on thermal data, but were empirically verified on our approach.

\vspace{-2ex}
\paragraph{Viewing angle.} 
Prior studies (like \cite{seits2022role}) show that near-nadir views maximize canopy penetration, while oblique angles mainly improve SfM stability rather than foliage perforation.

\vspace{-2ex}
\paragraph{Sampling density.} 
Measuring sampling density is done by counting pixel- or camera rays per unit area.
Higher image density increases the likelihood of capturing ground pixels through transient canopy gaps, but also raises computational load. 
Previous thermal AOS work \cite{schedl2021autonomous} reports diminishing returns beyond 20--25 samples; we empirically validate this trend for RGB data in Section~\ref{sec:detection}.

\vspace{-2ex}
\paragraph{Ground sample distance (GSD).} 
A finer GSD improves separation between canopy and ground but increases total pixel count and processing time. 
We hypothesize that a GSD of approximately 1--2~cm offers a good trade-off, consistent with the public AOS dataset \cite{schedl2021airborne}.

\begin{figure}[t]
    \centering
    \includegraphics[width=0.45\linewidth]{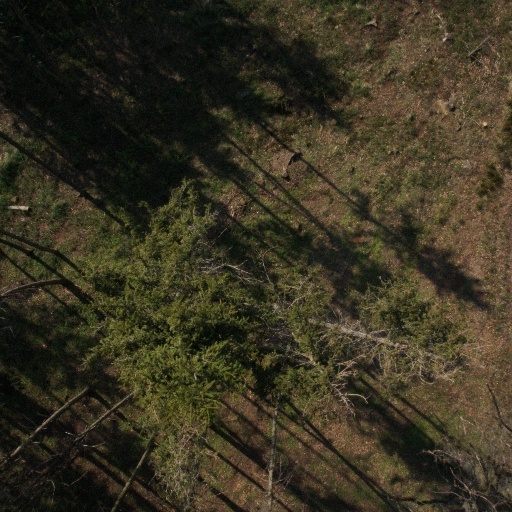} \,
    \includegraphics[width=0.45\linewidth]{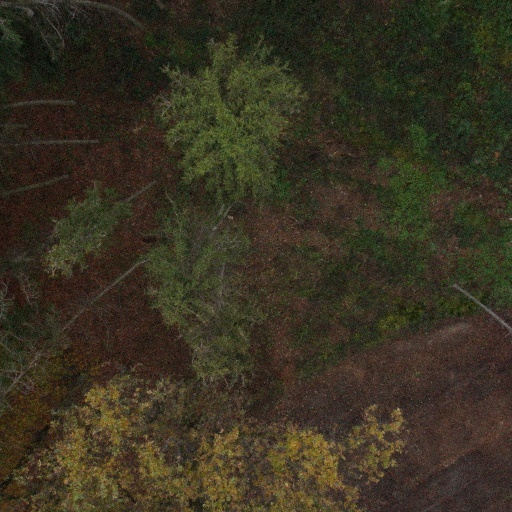} \\
    \includegraphics[width=0.95\linewidth]{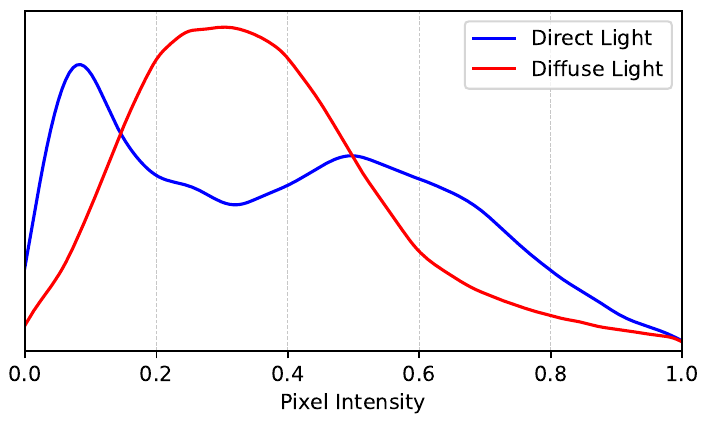}
    \caption{
        \textbf{Effect of lighting conditions}
        on the result on two sample views of the same location in different times, from \cite{schedl2021airborne}. 
        Direct sunlight (F11, top left) creates cast shadows, which might cause over exposed canopy and/or underexposed understory, as opposed to indirect lighting (F6, top right).
        This is also visible in the histogram (bottom) where even after equalization, a big portion of the dynamic range in direct lighting is spent on the lit area, and make the shadows more sensitive to quantization.
        See Figure~\ref{fig:loss} for more results.
        \label{fig:lighting}
    }
    \vspace{-3ex}
\end{figure}

\vspace{-2ex}
\paragraph{Lighting conditions.} 
Unlike prior work, we explicitly analyze the effect of illumination. 
Both thermal and RGB sensors adjust exposure to optimize dynamic range, but for RGB-based reconstruction, lighting plays a critical role. 
Our method performs best under diffuse light (overcast or twilight), which balances canopy and ground exposure. 
In contrast, direct sunlight creates harsh shadows and high contrast, forcing a trade-off between overexposed foliage and underexposed ground, often leaving parts of the understory unrecoverable. 
Figure~\ref{fig:lighting} illustrates this: diffuse illumination yields a unimodal brightness histogram, while direct light produces a bimodal distribution dominated by high-intensity pixels. Overall, diffuse light enhances both SfM stability and ground detail recovery. We hypothesize that similar effects may also apply to thermal imaging.

\subsection{Relative camera pose estimation.}
NeRF~\cite{mildenhall2020nerf}, as well as most reconstruction methods, requires relative camera pose for each image in the dataset. 
For the case of canopy penetration the need for  accurate reconstruction is especially important, as some of the occluding objects like branches and leaves are not much larger than a single pixel, if at all.
In our method, we follow the standard SfM pipeline of 2D feature extraction, feature matching (including outlier rejection), and bundle adjustment, which is a nonlinear optimization of camera parameters and poses, as well as 3D position of tie points.

\subsection{3D reconstruction using NeRF}\label{sec:nerf}
We describe specific details of the NeRF paradigm that are of importance to under canopy imaging, and leave specific implementation choices to Section~\ref{sec:experiments} below.

As mentioned, NeRF \cite{mildenhall2020nerf} defines an MLP which accepts a position in 3D $\mathbf{x}=(x,y,z)$ and 2D view direction $\mathbf{d}=(\phi,\theta)$, and outputs $\mathbf{c}=(r,g,b)$ and $\sigma$.
The expected color $C(\mathbf{r})$ of camera ray $\mathbf{r}(t)=\mathbf{o} + t\mathbf{d}$ with near and far bounds $t_1$ and $t_2$ is
\begin{equation} \label{eq:nerf}
  \hat C(\mathbf r)=\int_{t_1}^{t_2}T(t)\,\sigma(t)\,\mathbf{c}(t,\mathbf d)\,dt,
\end{equation}
where \(T(t)=\exp\bigl(-\int_{t_1}^t\sigma(s)\,ds\bigr)\) is the transmission function.
During training of the MLP, we assume camera position and intrinsics are known, and try to reconstruct given images pixels $C(\mathbf r)$ with the output of \eqref{eq:nerf}, typically with the L1 loss
\begin{equation} \label{eq:l1loss}
    L_{1}(\hat C, C) = \sum_{p \in I} |\hat C(\mathbf r_p)- C(\mathbf r_p)| , 
\end{equation}
for each ray $r_p$ corresponding to each pixel $p$ in each image $I$ of the dataset.

\paragraph{Low light loss.} \label{sec:rawloss}
To better cope with poorly lit parts of the understory, we adopt the loss function presented by Mildenhall \etal~\cite{mildenhall2022rawnerf}, which uses approximated tonemapping of the form
\begin{equation} \label{eq:rawloss}
    L_{raw}(\hat C, C) = \sum_{p \in I} \left(\frac{\hat C(\mathbf r_p)- C(\mathbf r_p)}{\mathrm{sg}(\hat C(\mathbf r_p)) + \epsilon}\right)^2 , 
\end{equation}
where `$\mathrm{sg}$' indicates \textit{stop gradient}, and $\epsilon$ is a tolerance value empirically set to be $10^{-3}$.
This loss can be used instead-, or in addition to- \eqref{eq:l1loss}, and puts an emphasis on low light pixels, and hence is better fitted for high dynamic range (HDR) forest scenes.

\subsection{Ground only Rendering} \label{sec:peal}
We describe two methods to generate an image of the understory and the ground, based on the trained NeRF model

\paragraph{Cropping out the canopy by terrain height.}
As mentioned, one straightforward way to reveal terrain, is to simply modify \eqref{eq:nerf} and start the integration slightly above ground height $t_{g}$
\begin{equation}\label{eq:crop}
  \hat C_{crop}(\mathbf r)=\int_{t_{g}}^{t_2}T(t)\,\sigma(t)\,c(t,\mathbf d)\,dt.
\end{equation}
Following prior works like~\cite{schedl2021airborne,thiel2020uas}, this height can be derived from a DTM obtained beforehand, coupled with the drone telemetry data.

\paragraph{Canopy masking using segmentation.}
The previous method operated directly in the 3D scene, but an alternative approach exists, which extracts more data from the 2D images beforehand.
Specifically, we aim to train the NeRF's MLP with an additional, fifth\footnote{in addition to 3 colors (RGB) and the density $\sigma$}, output, indicating the presence of occluding canopy.
We denote this additional information by $v(t)$, and following works like \cite{mirzaei2023spin,yin2023or} we use it to mask away the canopy by incorporating it into \eqref{eq:nerf} in a manner similar to the transmission function $T$,
\begin{equation} \label{eq:seg}
  \hat C_{seg}(\mathbf r)=\int_{t_1}^{t_2}T(t)\,\sigma(t)\,v(t)\,c(t,\mathbf d)\,dt.
\end{equation}
During training, $v(t)$ is optimized exactly like one of the color outputs RGB, but using a 2D segmentation map as supervision.
As mentioned earlier, this 2D segmentation map can be obtained by state of the art segmentation models, even with little to no extra supervision~\cite{fan2022nerf}.
Interestingly, however, we find that sometimes a simple color based segmentation works surprisingly well, when the ground color is different from the canopy. 
Given the color of a single pixel selected by the user, we use a box region in HSV space to flag the canopy.

\section{Experiments}
\label{sec:experiments}
We designed three experiments to evaluate the effectiveness and robustness of our under-canopy reconstruction pipeline across different challenges. 
First, we conduct a comparison of multiple reconstruction strategies. 
Next, we evaluate the downstream task of person detection under foliage.
Finally, we examine its applicability to forest inventory by testing tree stem detection on 3D models with the canopy removed.

\paragraph{Implementation Details.}
We used COLMAP~\cite{schonberger2018robust} to estimate (relative) camera poses.
The 3D scene reconstruction was based on the Instant-NGP implementation of NeRF~\cite{mueller2022instant}, with modifications to the loss function and segmentation peeling, as described in Section~\ref{sec:rawloss} and~\ref{sec:peal} respectively.
Additionally, we added a dynamic memory manager to handle the large amount of images.
We also adjusted several Instant-NGP hyperparameters from their default values: Batch size $\sim 2\cdot10^6$, hash map of $23$ bits, base resolution $32$.
Since NeRF is computationally demanding, the entire pipeline was executed on a laptop equipped with a 16-core Intel i9 CPU, 32~GB RAM, and an NVIDIA RTX~4090 GPU.
All runs were completed in under 25 minutes, of which approximately 20 were dedicated to NeRF training.

\paragraph{Datasets.}
All experiments except one (see below) were conducted on the publicly available dataset from the AOS paper~\cite{schedl2021airborne}, which includes both visible-spectrum RGB and thermal LWIR imagery from a drone.
While camera positions are provided with the dataset, we re-estimated them from scratch using COLMAP~\cite{schonberger2018robust} to allow feeding the original, non-rectified, full-resolution images into NeRF~\cite{mueller2022instant}, thereby avoiding data loss.
The final dataset used in our experiments comprises 11 out of the 12 available 'F' scenes\footnote{we do not include the 'O' flights, which were used for training in the original paper~\cite{schedl2021airborne}, as they contain no occlusion}, as SfM reconstruction failed for scene F7, likely due to its extremely low-light conditions.
Additional details on the SfM results are provided in the \suppmat{}, as well as the resulting NeRF rendering.
We further discuss the visual quality of the NeRF under-canopy images in Section~\ref{sec:detection}, as part of the person detection task.

We additionally use a synthetic CAD 3D scene of a dense forest using blender~\cite{blender2018}, which unlike \cite{schedl2021airborne} contains exact ground-truth surface image.
For more details on the creation of the synthetic data please see the \suppmat{}.

\subsection{Ground surface reconstruction quality}\label{sec:quality}

We conducted three evaluations to assess the reconstruction quality of our approach.
First, a comparison of our method against other reconstruction techniques that operate on RGB data.
Second, a comparison of different flavors of the proposed method, especially the choice of loss function and segmentation.
Lastly, we use synthetic data to get a qualitative measurement of the surface reconstruction results.

\subsubsection{Reconstruction quality on RGB}
To evaluate the reconstruction quality of our approach on RGB data, we compared four methods using the F6 flight from the AOS dataset~\cite{schedl2021airborne}.
The qualitative results are shown in Figure~\ref{fig:teaser} and summarized below:

{\setlength{\parskip}{-13pt}
    \paragraph{AOS method~\cite{schedl2021airborne}.} 
    This method was originally designed to run on thermal, rather than RGB data. 
    Indeed, once the rays are refocused to ground level, the result is blurred and no detail can be retrieved.
    
    \paragraph{Multi-View Stereo (MVS).} We used the implementation provided in COLMAP~\cite{schonberger2018robust} with default parameters, followed by cropping the canopy by DTM in a manner similar to Section~\ref{sec:peal}. 
    This method assumes smoothness of the 3D scene, and therefore fails to capture 3D elements under the canopy, which is mostly of complex structure. 
    For more details on the smoothness assumption, see section 4.3 in~\cite{schoenberger2016mvs}. 
    
    \paragraph{3D Gaussian Splatting (3DGS)~\cite{kerbl20233d}.}
    The recent open source implementation from ~\cite{lichtfeld2025} outperforms MVS, but still some parts of the scene are not reconstructed.
    Even though this method is similar to NeRF (in the sense that the optimization is appearance based), there are some fundamental differences that might cause the failure.
    Key limitations include: (i) initialization from the sparse SfM point cloud, which typically does not contain any data in the occluded under-canopy regions; (ii) lack of lens distortion handling; and (iii) rasterization-based depth sampling, which introduces artifacts such as the "bouncing" effect~\cite{radl2024stopthepop}.
        
    \paragraph{Our method.} Based on Instant-NGP~\cite{mueller2022instant} with custom modifications, our approach achieves the most complete and visually coherent reconstruction among all tested methods.
    Unlike MVS or 3DGS, which rely on explicit geometry or dense point initialization, NeRF optimizes a continuous volumetric representation and integrates color and density along camera rays. 
    This allows the model to exploit partial visibility through canopy gaps and infer plausible structure in occluded regions, with no extra assumptions.
    Combined with low-light loss function, our method is particularly effective for under-canopy reconstruction.    
}

\begin{figure}[b]
  \centering
  \setlength{\tabcolsep}{3pt}
  \begin{tabular}{rcc}
    &
    L1 loss~\eqref{eq:l1loss} &
    RAW loss~\eqref{eq:rawloss} \\
    \rotatebox{90}{\hspace{9ex} F11} &
    \begin{tikzpicture}
        \node[anchor=south west,inner sep=0] (image) at (0,0) {\includegraphics[width=0.45\linewidth]{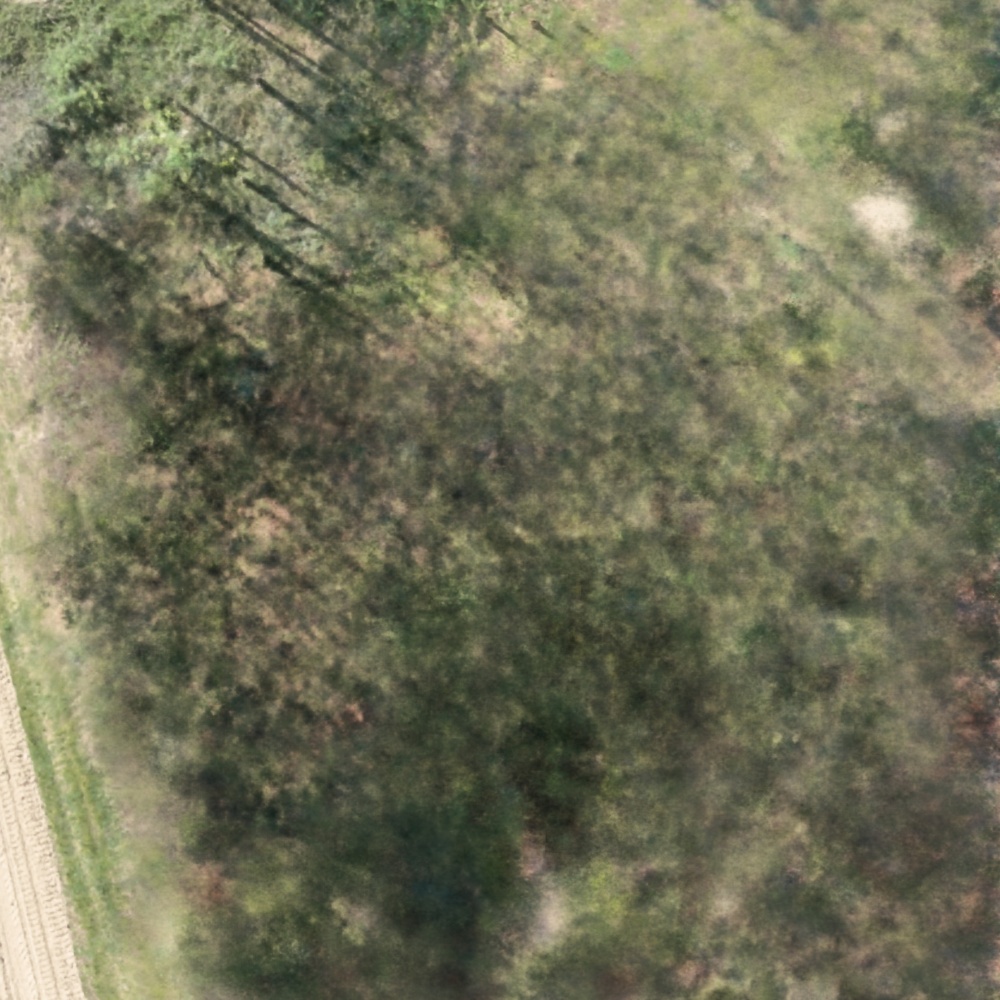}};
        \begin{scope}[x={(image.south east)},y={(image.north west)}]
            \draw[red,ultra thick,rounded corners] (.5,.75) rectangle (.95,.99);
            \draw[red,ultra thick,rounded corners] (.05,.05) rectangle (.2,.4);
        \end{scope}
    \end{tikzpicture} &
    \begin{tikzpicture}
        \node[anchor=south west,inner sep=0] (image) at (0,0) {\includegraphics[width=0.45\linewidth]{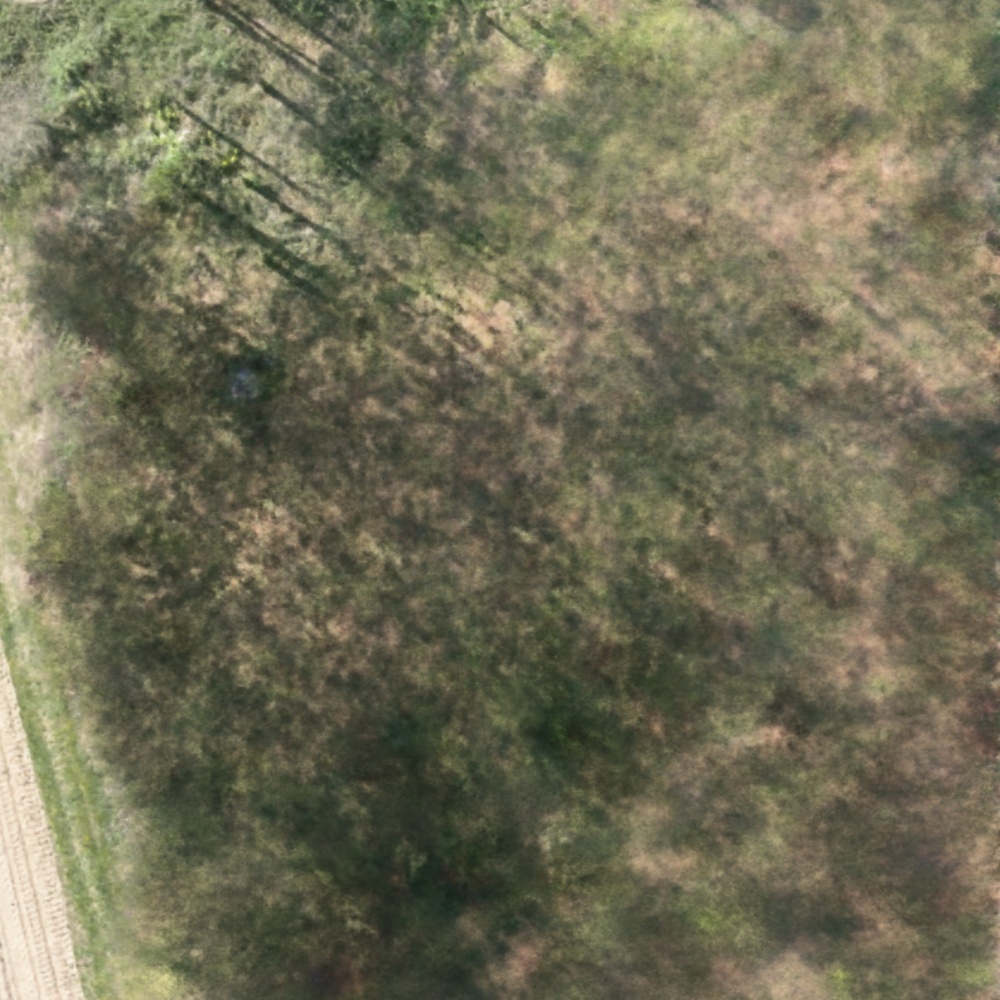}};
        \begin{scope}[x={(image.south east)},y={(image.north west)}]
            \draw[red,ultra thick,rounded corners] (.5,.75) rectangle (.95,.99);
            \draw[red,ultra thick,rounded corners] (.05,.05) rectangle (.2,.4);
        \end{scope}
    \end{tikzpicture} \\
    \rotatebox{90}{\hspace{9ex} F5} &
    \begin{tikzpicture}
        \node[anchor=south west,inner sep=0] (image) at (0,0) {\includegraphics[width=0.45\linewidth]{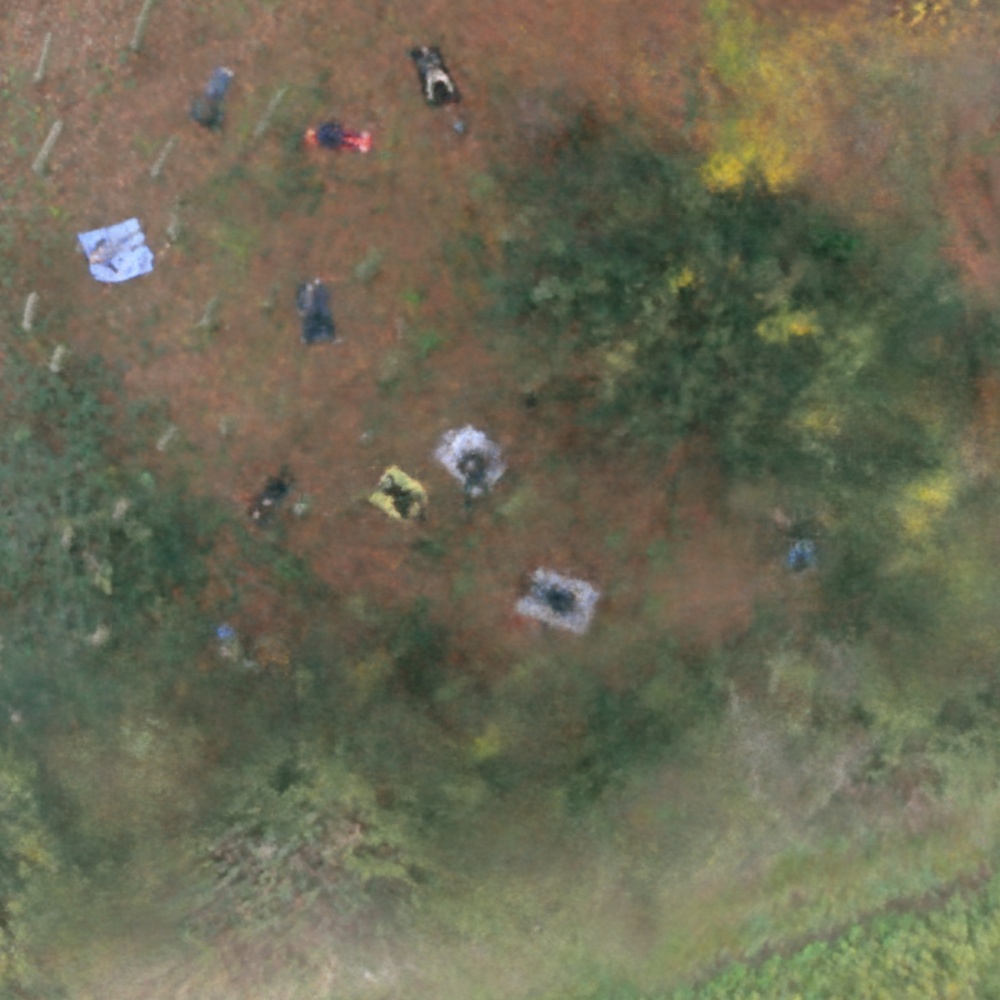}};
        \begin{scope}[x={(image.south east)},y={(image.north west)}]
            \draw[red,ultra thick,rounded corners] (.35,.35) rectangle (.62,.60);
            \draw[red,ultra thick,rounded corners] (.749,.39) rectangle (.851,.51);
        \end{scope}
    \end{tikzpicture} &
    \begin{tikzpicture}
        \node[anchor=south west,inner sep=0] (image) at (0,0) {\includegraphics[width=0.45\linewidth]{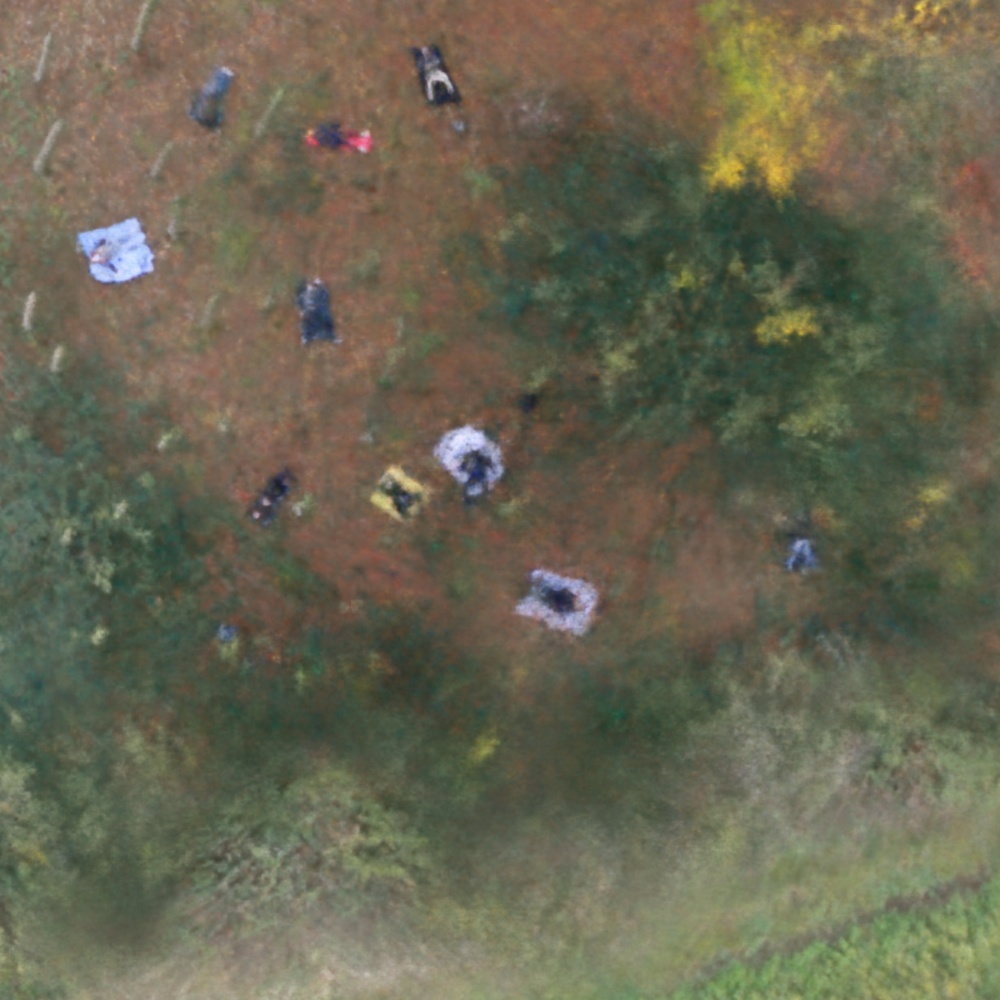}};
        \begin{scope}[x={(image.south east)},y={(image.north west)}]
            \draw[red,ultra thick,rounded corners] (.35,.35) rectangle (.62,.60);
            \draw[red,ultra thick,rounded corners] (.749,.39) rectangle (.851,.51);
        \end{scope}
    \end{tikzpicture} \\
  \end{tabular}
  \caption{
  \textbf{A comparison of two NeRF loss functions,}
  L1~\eqref{eq:l1loss} and RAW~\eqref{eq:rawloss},on two scenes from \cite{schedl2021airborne}.
  The two images top are from the F11 scene, which includes direct lighting and high dynamic range, the L1 loss struggles with the cast shadows and shade boundaries.
  The two bottom images are from the F5 scene, where dynamic range is more balanced, and RAW loss better captures the most occluded person on the right part of the image, which can be partially seen in the L1 image.
  Additionally, other persons are slightly more blurred in the L1 image.
  See the \suppmat{} for results on more loss functions. 
  \label{fig:loss}    
  }
  \vspace{-4ex}
\end{figure}

\subsubsection{Flavors of the proposed method}
We conducted a qualitative ablation study to analyze two key components of our method, \ie loss function and the canopy-removal strategy. 
This evaluation was performed on two scenes from the AOS dataset~\cite{schedl2021airborne}, using different configurations.

\paragraph{Loss function.} 
As discussed in Section~\ref{sec:rawloss}, we advocate the usage of RAW loss~\eqref{eq:rawloss} for foliage removal.
Most NeRF implementations optimize either an L1 loss~\eqref{eq:l1loss}, L2 loss, or the Huber loss~\cite{huber1992robust}, which is a combination of the two. 
We observe that for the case of foliage removal either L1 or RAW loss yields the highest quality results, as depicted in Figure~\ref{fig:loss} (we provide results for other loss functions in the \suppmat{}).
RAW-NeRF achieved the most consistent coherent results especially in scenes with large dynamic range, where both dark and bright pixels need to be reconstructed. 
Additionally, even in more balanced scenes such as F5, a person in the right side of the image is clearly more visible and other persons seem sharper.

\begin{figure}[b]
  \centering
  \small
  \setlength{\tabcolsep}{5pt}
  \begin{tabular}{cc}
    \includegraphics[width=0.45\linewidth]{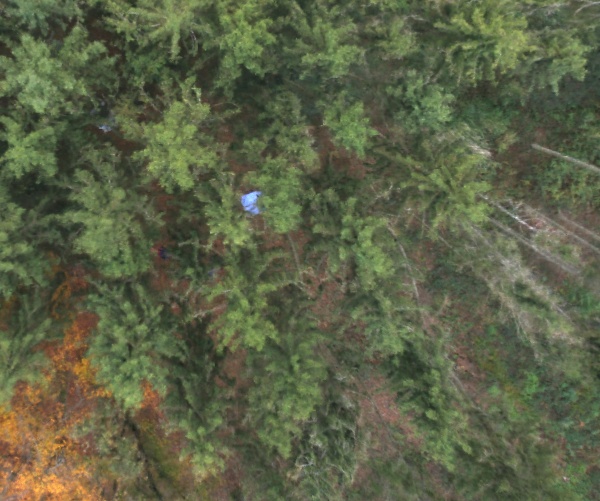} &
    \includegraphics[width=0.45\linewidth]{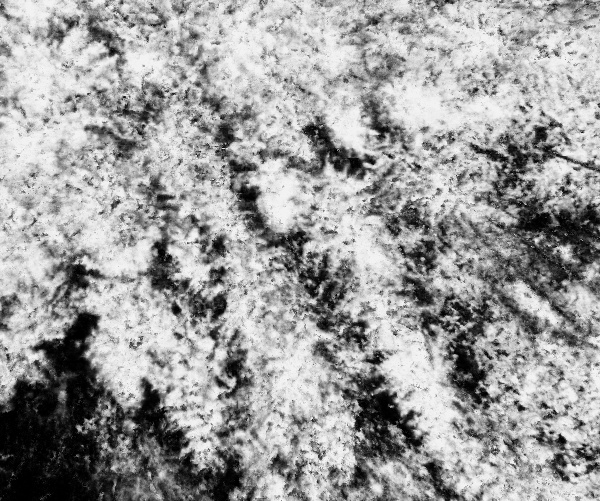} \\
    Full scene &
    3D color segmentation mask \\
    \includegraphics[width=0.45\linewidth]{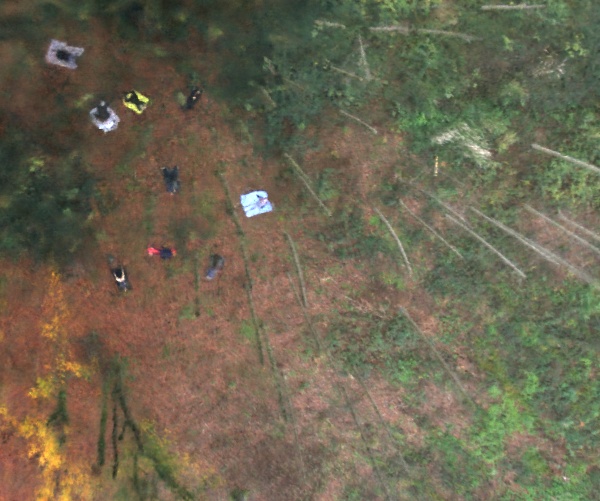} &
    \includegraphics[width=0.45\linewidth]{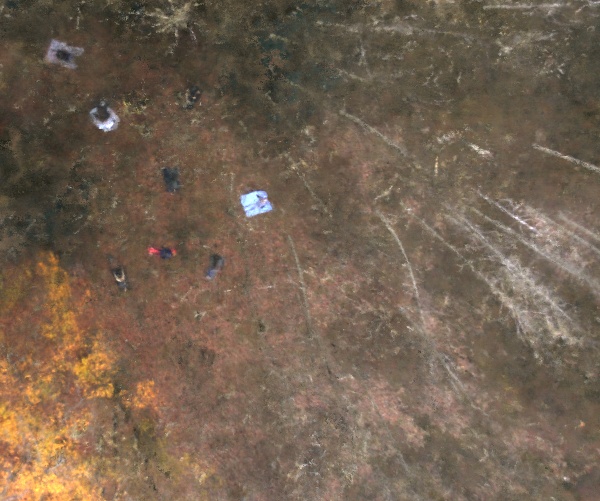} \\
    3D Crop~\eqref{eq:crop} &
    Segmentation~\eqref{eq:seg} \\
  \end{tabular}
  \caption{
  \textbf{A Comparison of canopy removal methods.}
  The two methods described in Section \ref{sec:peal}, are applied on scene F5 from \cite{schedl2021airborne}.   
  The full scene (before canopy removal) is in the top left, 3D crop bottom left, and 3D segmentation bottom right.
  In addition, the 3D color-based segmentation mask is rendered as grayscale image, where \eg the trunks in the right side are excluded.
  We recommend the reader to view the \suppvideo{supp_Figure5.mp4} of this segmentation result.
  See Section~\ref{sec:peal} for more details.  
  \label{fig:peel}    
  }
  \vspace{-2ex}
\end{figure}

\paragraph{Canopy removal method.} 
As mentioned in Section~\ref{sec:peal}, we propose two approaches for removing the canopy, \ie height-based cropping~\eqref{eq:crop} and segmentation-based filtering~\eqref{eq:seg}.
Figure~\ref{fig:peel} illustrates both approaches on the F5 scene, alongside the original reconstruction (with the canopy).
The segmentation method relies on color distance from a canopy pixel selected by the user.
While effective at removing foliage and green shrubs, it also eliminates nearby green-tinted objects, such as trunks with moss, and even a person lying on a yellow-green mattress.
Interestingly, trunks with gray hue are kept completely, including some fine branches.
The height cropping approach avoids these phenomena, but requires careful tuning of the height threshold $t_{g}$ to maximize canopy removal while preserving ground details. 
This latter tradeoff is influenced by potential inaccuracies in the DTM and drone telemetry.

Unless stated otherwise, we use the 3D crop approach throughout this paper.

\subsubsection{Synthetic data quantitative results.} 
In order to gain some qualitative comparison between 3DGS~\cite{lichtfeld2025} and our method, we use the aforementioned synthetic data, which provides ground-truth image of the surface, to which we can compare.

\begin{figure}[ht]
    \centering
    \setlength{\tabcolsep}{5pt}
    \begin{tabular}{cc}
        \includegraphics[width=0.43\linewidth]{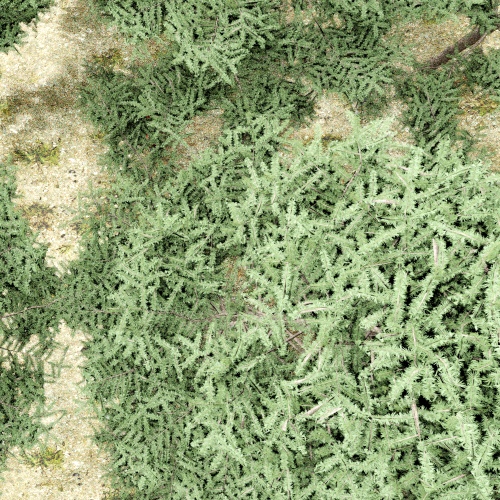} &
        \includegraphics[width=0.43\linewidth]{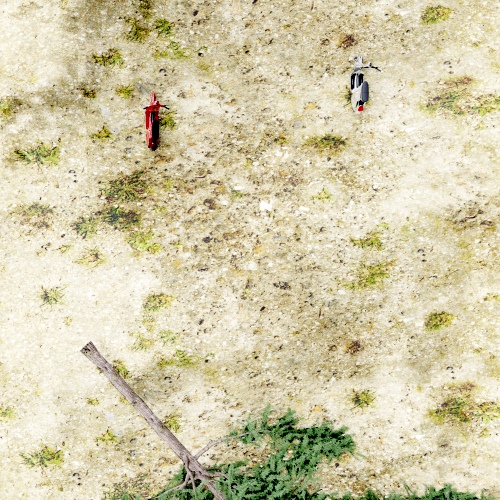} \\
        Scene w/ canopy &
        Scene w/o canopy \\
        \includegraphics[width=0.43\linewidth]{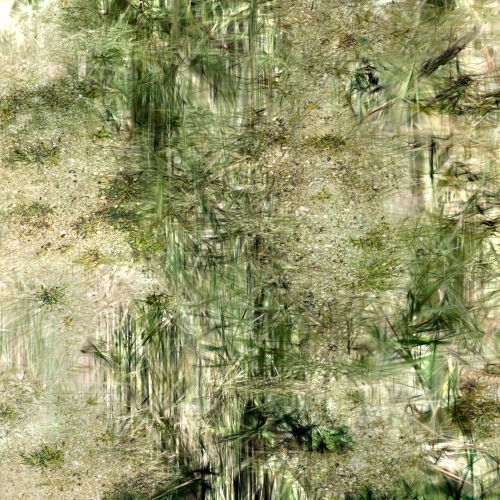} &
        \includegraphics[width=0.43\linewidth]{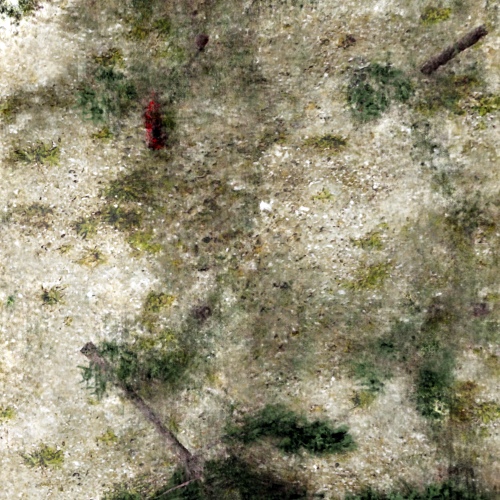} \\
        3DGS~\cite{lichtfeld2025} &
        Our method \\
    \end{tabular}
    \caption{
        Sample synthetic image and qualitative results on synthetic data.
        See the \suppmat{} for more details.
    }
    \label{fig:sim_vis}
    \vspace{-1ex}        
\end{figure}

Table~\ref{tab:sim_ssim} report M-SSIM~\cite{wang2003multiscale} score to measure the reconstruction quality of each of the methods, both with- and without the canopy (using 3D crop). 
While both methods are comparable for the full scene, 3DGS falls short when it comes to the surface reconstruction.
This can also be seen in Figure~\ref{fig:sim_vis}, where 3DGS contains much more artifacts.

\begin{table}[h!]
    \centering
    \begin{tabular}{ccc}
        \hline
        \hline
        Type & w/ canopy & w/o canopy \\
        \hline
        3DGS~\cite{lichtfeld2025} & $0.64\pm 0.013$ & $0.40 \pm 0.044$ \\
        Ours                      & $0.65\pm 0.031$ & $0.44 \pm 0.042$ \\
        \hline
        \hline
    \end{tabular}    
    \caption{
        M-SSIM~\cite{wang2003multiscale} (higher is better) reconstruction quality of two methods on  a synthetic 3D scene. 
        Presented are mean$\pm$std over several views of the scene.
        See text for details
    }
    \label{tab:sim_ssim}
    \vspace{-1ex}    
\end{table}

\subsection{Person Detection for SAR.}\label{sec:detection}

We selected a two-stage detection strategy that requires \textbf{no SAR-specific annotations}, making it suitable for scenarios where labeled data is scarce. 
First, we run a COCO-pretrained \texttt{yolo11n}~\cite{yolo11_ultralytics} detector with relaxed thresholds (confidence 0.01, NMS 0.15), which was empirically verified to generate \textbf{high-recall} region proposals from ground-only mosaics. 
Second, the proposals undergo unsupervised anomaly filtering based on the reconstruction error of an autoencoder trained on random background patches of under-canopy images that do not contain human targets.
More details on the architecture and training regime of the autoencoder are provided in the \suppmat{}. 
This design requires no SAR-specific supervision and leverages the distinct appearance of human shapes in foliage-free renderings.

Importantly, not all persons were uncovered in scenes F1, F3, and F4 (see \suppmat{} for details), and we present performance metrics only for persons that were at least partially revealed in the reconstruction, as was done in the original dataset~\cite{schedl2021airborne}.
We hypothesize this may be due to suboptimal dynamic range, but mainly due to the structural complexity of broadleaf or mixed canopies, which create dense occlusions. 
Higher spatial resolution and improved exposure control could mitigate these issues, and even thermal AOS imagery for these scenes shows reduced quality, suggesting that these conditions challenge multiple sensing modalities.

On the AOS dataset~\cite{schedl2021airborne}, our method achieved {80.5\% AP} (18 FP, 34 TP), improving to {89.6\% AP} when excluding two low-quality scenes (F3, F4), as shown in Table~\ref{tab:atr}.
While thermal AOS remains superior (92.2\% AP), our RGB-only pipeline demonstrates promising performance without specialized sensors. 
Our method is also attractive compared to HSV-based AOS~\cite{ryu2023enhanced}, which also operates on RGB, even though it was tested on a disjoint synthetic data that was not made public. 
While intentionally simple, our design demonstrates the feasibility of SAR detection from foliage-free reconstructions and leaves substantial room for future improvements.

\begin{table}[h]
    \centering
    \begin{tabular}{ccccc}
         \hline
         Method & Image type & AP & FP & TP \\
         \hline 
         \hline
         AOS~\cite{schedl2021airborne} & Thermal &92.2\% & 2 & 52 \\  
         AOS~\cite{ryu2023enhanced}  & HSV & 64.9\%$^{\ddagger}$ & \small{N/A} $^{\dagger}$ &  \small{N/A} $^{\dagger}$\\  
         \hline
         Our result & RGB & 80.5\% & 18 & 34 \\
         Our result $^{\star}$ & RGB & 89.6\% & 10 & 33\\
         \hline
    \end{tabular}
    \caption{
    \textbf{Person detection performance} on the `F' scenes from \cite{schedl2021airborne}.
    $^{\ddagger}$ Averaged over multi seasonal data.
    $^{\dagger}$ Ryu \etal~\cite{ryu2023enhanced} used disjoint synthetic dataset, so only percentage is included.
    $^{\star}$ excluding F3 and F4.
    }
    \label{tab:atr}
\end{table}

Sensitivity analysis (Table~\ref{tab:partial}) reveals more details regarding the robustness of the method. 
removing 25\% of the images causes a drop in performance, mainly in the more dense forests (F0-F4), while on conifer scenes (F5-F6) performance remains stable down to approximately 66\% of the original image. 
False alarms count also rises with the reduction of images, which can probably be mitigated with further tuning of the anomaly detector.
These results leave room for more in depth analysis of the image acquisition details, especially in more dense forests.

\begin{table}[t]
    \centering
    \begin{tabular}{lrrcc}
         \hline
          Sampling     & \% of images & AP & FP & TP  \\
         \hline 
         \hline
          All          & 100\% & 80\% & 18 & 34 \\
          Drop every 4 & 75\%  & 35\% & 47 & 20 \\
          Drop every 3 & 66\%  & 31\% & 54 & 17 \\
          Drop every 2 & 50\%  & 34\% & 48 & 20 \\
          Keep every 3 & 33\%  & 27\% & 46 & 7 \\
         \hline         
    \end{tabular}
    \caption{
    \textbf{Sample density sensitivity} of our method on the `F' scenes from \cite{schedl2021airborne}.
    }
    \label{tab:partial}
    \vspace{-2ex}
\end{table}

\begin{figure}[b]
    \centering 
    \setlength{\tabcolsep}{1pt}
    \begin{tabular}{ccc}
        \includegraphics[width=0.33\linewidth]{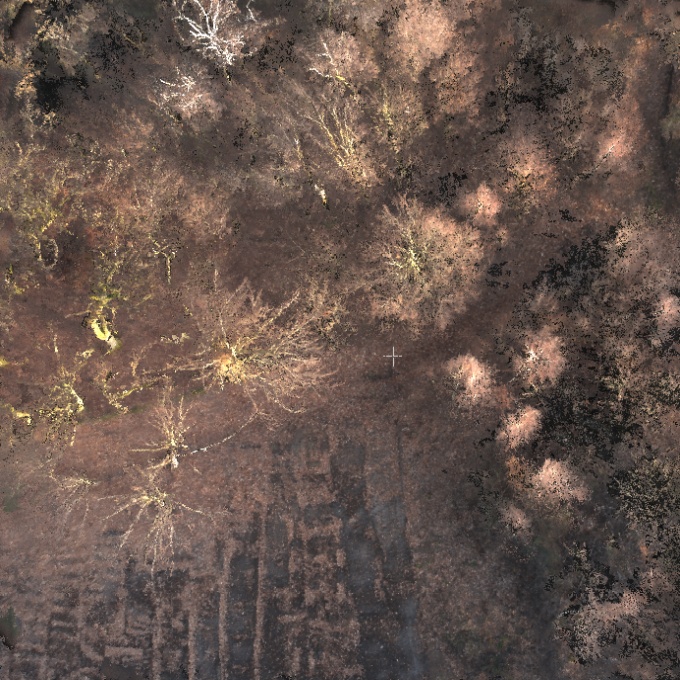} &
        \includegraphics[width=0.33\linewidth]{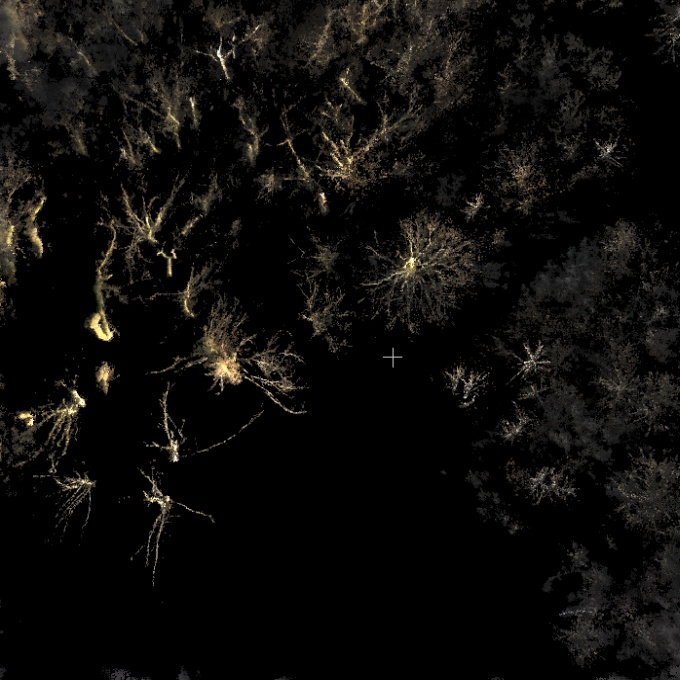} &
        \includegraphics[width=0.33\linewidth]{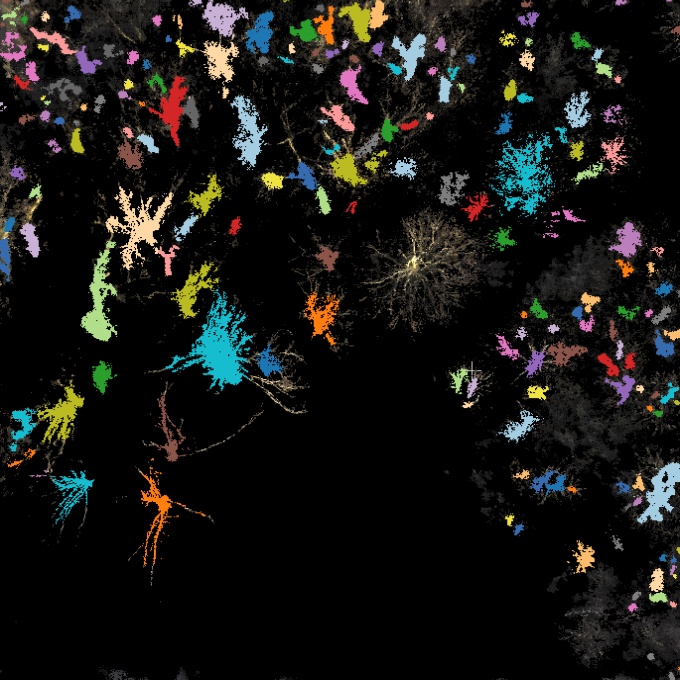} \\
        Full scene &
        \parbox{0.3\linewidth}{\centering Canopy \& ground removed} &
        \parbox{0.3\linewidth}{\centering HDBSCAN clusters} \\
    \end{tabular}
    \caption{Stages of point cloud processing and stem clustering on the F9 scene from the AOS dataset~\cite{schedl2021airborne}.}
    \label{fig:stems_peeling}
    \vspace{-3ex}
\end{figure}

\subsection{Tree stem counting}
The last experiment aims to illustrate the utility of the foliage-free reconstructions for a downstream task such as forest inventory.
We export the trained NeRF to a 3D point cloud by sampling the density field~$\sigma$ over a dense regular grid, retaining only points with sufficient density. 
We apply a three-step post-processing pipeline to obtain a foliage-free point cloud dominated by vertical stem fragments:
First, we then apply color-based foliage removal using segmentation, followed by a 3D crop that prunes remaining high-canopy branches, and lastly a similar crop from below to remove terrain by retaining only points whose height is above the local DTM.
On this processed cloud, we run HDBSCAN~\cite{mcinnes2017hdbscan} to cluster points into stem candidates and discard clusters whose volumetric extent is too small or too large.
We additionally estimate the tilt of each stem cluster using PCA, which allows us to filter out non-vertical candidates and merge clusters that likely belong to the same tree when one cluster appears directly above another.
We selected this clustering approach because it is nearly parameter-free and operates in a bottom-up manner, allowing each stem cluster to form adaptively while maintaining separation from neighboring clusters.
Figure~\ref{fig:stems_peeling} illustrates these stages on the F9 scene from the AOS dataset~\cite{schedl2021airborne}, showing the full scene, the processed cloud, and the clustering result with random cluster coloring.
Most stems appear well isolated, providing a reasonable estimate of tree count and typical height.
We recommend the reader to view the \suppvideo{supp_Figure6.mp4} of this result.

\section{Conclusions}

We presented a NeRF-based 3D reconstruction pipeline capable of revealing terrain beneath dense forest canopies using only airborne RGB imagery.
This approach is a cost effective alternative to existing approaches that require specialized sensors like LiDAR and thermal camera.
To highlight the quality and versatility of these reconstructions, we applied a set of intentionally simple (even naive) analytics: color based segmentation, basic stem clustering, and unsupervised person detection.
These were chosen deliberately to emphasize the inherent quality of the reconstructions and to leave substantial room for improvement through more advanced methods.

Future work will explore ways to extend the NeRF module in our approach, by a combination with other successful ones such as
multi spectral fusion~\cite{poggi2022xsrnf}, online reconstruction~\cite{rosinol2022nerf} and modern implementations of 3DGS like~\cite{wu20253dgut,moenne20243d}.
In addition, we aim to leverage concepts from forest monitoring literature, such as Leaf Area Index (LAI)~\cite{chen1993defining}, which may serve as a predictive indicator for canopy penetration success.

\vspace{-2ex}
\paragraph{Limitations and Failure Cases.}
While our pipeline demonstrates promising results, several environmental and technical factors limit its effectiveness.
First, performance strongly depends on ambient lighting conditions, unlike LiDAR and thermal imaging which rely on active illumination or radiant objects, respectively.
This was especially visible in scenes F1, F3 and F4 where the canopy occlusion was more severe.
We therefore see potential in using our method with either thermal or low light\footnote{for their high dynamic range} cameras in these cases.
Second, reconstruction quality is sensitive to the accuracy of the SfM solution, which can degrade under dynamic foliage or poor image quality. This may be solved using motion-aware SfM and improved telemetry integration.

Despite these challenges, our method retains strong performance across a range of conditions, and ongoing improvements in acquisition planning and model robustness can further extend its applicability.


\small
\begin{thebibliography}{10}\itemsep=-1pt

\bibitem{akeley2002will}
Kurt Akeley, David Kirk, Larry Seiler, Philipp Slusallek, and Brad Grantham.
\newblock When will ray-tracing replace rasterization?
\newblock In {\em ACM SIGGRAPH 2002 conference abstracts and applications}, pages 86--87, 2002.

\bibitem{axelsson2000dem}
Peter Axelsson.
\newblock Dem generation from laser scanner data using adaptive tin models.
\newblock In {\em Int. Archives of Photogrammetry and Remote Sensing}, volume~33, pages 110--117, 2000.

\bibitem{campbell2018quantifying}
Michael~J Campbell, Philip~E Dennison, Andrew~T Hudak, Lucy~M Parham, and Bret~W Butler.
\newblock Quantifying understory vegetation density using small-footprint airborne lidar.
\newblock {\em Remote sensing of environment}, 215:330--342, 2018.

\bibitem{caulfield2018whats}
Brian Caulfield.
\newblock What’s the difference between ray tracing and rasterization?
\newblock \url{https://blogs.nvidia.com/blog/whats-difference-between-ray-tracing-rasterization}, 2018.

\bibitem{chase2011airborne}
Arlen~F Chase, Diane~Z Chase, John~F Weishampel, Jason~B Drake, Ramesh~L Shrestha, K~Clint Slatton, Jaime~J Awe, and William~E Carter.
\newblock Airborne lidar, archaeology, and the ancient maya landscape at caracol, belize.
\newblock {\em Journal of Archaeological Science}, 38(2):387--398, 2011.

\bibitem{chen1993defining}
Ji{\^a}n~M. Chen and T.~A. Black.
\newblock Defining leaf area index for non-flat leaves.
\newblock {\em Agricultural and Forest Meteorology}, 57(4):235--245, 1993.

\bibitem{chugunov2024neural}
Ilya Chugunov, David Shustin, Ruyu Yan, Chenyang Lei, and Felix Heide.
\newblock Neural spline fields for burst image fusion and layer separation.
\newblock In {\em Proceedings of the IEEE/CVF conference on computer vision and pattern recognition}, pages 25763--25773, 2024.

\bibitem{cmielewski2021uav}
Bart{\l}omiej {\'C}mielewski, Dominika Sieczkowska, Jacek Ko{\'s}ciuk, Jos{\'e}~M Bastante, and Izabela Wilczy{\'n}ska.
\newblock Uav lidar mapping in the historic sanctuary of machupicchu: challenges and preliminary results: Part 1.
\newblock {\em Wiadomo{\'s}ci Konserwatorskie}, 2021.

\bibitem{blender2018}
Blender~Online Community.
\newblock {\em Blender - a 3D modelling and rendering package}.
\newblock Blender Foundation, Stichting Blender Foundation, Amsterdam, 2018.

\bibitem{djistore}
DJI.
\newblock Dji store.
\newblock \url{https://store.dji.com/}.
\newblock Accessed: 2025-07-01.

\bibitem{drake2002estimation}
Jason~B Drake, Ralph~O Dubayah, David~B Clark, Robert~G Knox, J~Bryan Blair, Michelle~A Hofton, Robin~L Chazdon, John~F Weishampel, and Steve Prince.
\newblock Estimation of tropical forest structural characteristics using large-footprint lidar.
\newblock {\em Remote sensing of environment}, 79(2-3):305--319, 2002.

\bibitem{evans2013uncovering}
Damian~H Evans, Roland~J Fletcher, Christophe Pottier, Jean-Baptiste Chevance, Dominique Soutif, Boun~Suy Tan, Sokrithy Im, Darith Ea, Tina Tin, Samnang Kim, et~al.
\newblock Uncovering archaeological landscapes at angkor using lidar.
\newblock {\em Proceedings of the National Academy of Sciences}, 110(31):12595--12600, 2013.

\bibitem{fan2022nerf}
Zhiwen Fan, Peihao Wang, Yifan Jiang, Xinyu Gong, Dejia Xu, and Zhangyang Wang.
\newblock Nerf-sos: Any-view self-supervised object segmentation on complex scenes.
\newblock {\em arXiv preprint arXiv:2209.08776}, 2022.

\bibitem{huber1992robust}
Peter~J Huber.
\newblock Robust estimation of a location parameter.
\newblock In {\em Breakthroughs in statistics: Methodology and distribution}, pages 492--518. Springer, 1992.

\bibitem{yolo11_ultralytics}
Glenn Jocher and Jing Qiu.
\newblock Ultralytics yolo11.
\newblock \url{https://github.com/ultralytics/ultralytics}, 2024.

\bibitem{kerbl20233d}
Bernhard Kerbl, Georgios Kopanas, Thomas Leimk{\"u}hler, and George Drettakis.
\newblock 3d gaussian splatting for real-time radiance field rendering.
\newblock {\em ACM Trans. Graph.}, 42(4):139--1, 2023.

\bibitem{kirillov2023segment}
Alexander Kirillov, Eric Mintun, Nikhila Ravi, Hanzi Mao, Chloe Rolland, Laura Gustafson, Tete Xiao, Spencer Whitehead, Alexander~C Berg, Wan-Yen Lo, et~al.
\newblock Segment anything.
\newblock In {\em Proceedings of the IEEE/CVF international conference on computer vision}, pages 4015--4026, 2023.

\bibitem{kraus1998determination}
Karl Kraus and Norbert Pfeifer.
\newblock Determination of terrain models in wooded areas with airborne laser scanner data.
\newblock {\em ISPRS Journal of Photogrammetry and Remote Sensing}, 53(4):193--203, 1998.

\bibitem{li2021let}
Xiaoyu Li, Bo Zhang, Jing Liao, and Pedro~V Sander.
\newblock Let's see clearly: Contaminant artifact removal for moving cameras.
\newblock In {\em Proceedings of the IEEE/CVF International Conference on Computer Vision}, pages 2011--2020, 2021.

\bibitem{mcinnes2017hdbscan}
Leland McInnes, John Healy, and Steve Astels.
\newblock hdbscan: Hierarchical density based clustering.
\newblock {\em The Journal of Open Source Software}, 2(11):205, 2017.

\bibitem{mildenhall2022rawnerf}
Ben Mildenhall, Peter Hedman, Ricardo Martin-Brualla, Pratul~P. Srinivasan, and Jonathan~T. Barron.
\newblock {NeRF} in the dark: High dynamic range view synthesis from noisy raw images.
\newblock {\em CVPR}, 2022.

\bibitem{mildenhall2020nerf}
Ben Mildenhall, Pratul Srinivasan, Matthew Tancik, Jonathan~T. Barron, Ravi Ramamoorthi, and Ren Ng.
\newblock Nerf: Representing scenes as neural radiance fields for view synthesis.
\newblock In {\em ECCV}, 2020.

\bibitem{mirzaei2023spin}
Ashkan Mirzaei, Tristan Aumentado-Armstrong, Konstantinos~G Derpanis, Jonathan Kelly, Marcus~A Brubaker, Igor Gilitschenski, and Alex Levinshtein.
\newblock Spin-nerf: Multiview segmentation and perceptual inpainting with neural radiance fields.
\newblock In {\em Proceedings of the IEEE/CVF Conference on Computer Vision and Pattern Recognition}, pages 20669--20679, 2023.

\bibitem{moenne20243d}
Nicolas Moenne-Loccoz, Ashkan Mirzaei, Or Perel, Riccardo de Lutio, Janick Martinez~Esturo, Gavriel State, Sanja Fidler, Nicholas Sharp, and Zan Gojcic.
\newblock 3d gaussian ray tracing: Fast tracing of particle scenes.
\newblock {\em ACM Transactions on Graphics (TOG)}, 43(6):1--19, 2024.

\bibitem{mueller2022instant}
Thomas M{\"u}ller, Alex Evans, Christoph Schied, and Alexander Keller.
\newblock Instant neural graphics primitives with a multiresolution hash encoding.
\newblock In {\em ACM SIGGRAPH 2022 Conference Proceedings}, pages 1--15. ACM, 2022.

\bibitem{nathan2024reciprocal}
Rakesh John Amala~Arokia Nathan, Sigrid Strand, Dmitriy Shutin, and Oliver Bimber.
\newblock Reciprocal visibility for guided occlusion removal with drones.
\newblock {\em IEEE Geoscience and Remote Sensing Letters}, 2024.

\bibitem{pettorelli2013normalized}
Nathalie Pettorelli.
\newblock {\em The normalized difference vegetation index}.
\newblock Oxford University Press, USA, 2013.

\bibitem{poggi2022xsrnf}
Matteo Poggi, Fabio Tosi, and Stefano Mattoccia.
\newblock Xs-nerf: Cross-spectral neural radiance fields for sparse-view hyperspectral reconstruction.
\newblock In {\em ECCV}, 2022.

\bibitem{radl2024stopthepop}
Lukas Radl, Michael Steiner, Mathias Parger, Alexander Weinrauch, Bernhard Kerbl, and Markus Steinberger.
\newblock Stopthepop: Sorted gaussian splatting for view-consistent real-time rendering.
\newblock {\em ACM Transactions on Graphics (TOG)}, 43(4):1--17, 2024.

\bibitem{rosinol2022nerf}
Antoni Rosinol, John~J Leonard, and Luca Carlone.
\newblock Nerf-slam: Real-time dense monocular slam with neural radiance fields.
\newblock {\em arXiv preprint arXiv:2210.13641}, 2022.

\bibitem{rosu2020semi}
Radu~Alexandru Rosu, Jan Quenzel, and Sven Behnke.
\newblock Semi-supervised semantic mapping through label propagation with semantic texture meshes.
\newblock {\em International Journal of Computer Vision}, 128(5):1220--1238, 2020.

\bibitem{ryu2023enhanced}
KangSoo Ryu, Byungjin Lee, Dong-Gyun Kim, and Sangkyung Sung.
\newblock Enhanced airborne optical sectioning design via hsv color space for detecting human object under obscured aerial image environment.
\newblock {\em International Journal of Control, Automation and Systems}, 21(11):3734--3745, 2023.

\bibitem{schedl2020search}
Daniel Schedl, Michael Zangl, and Thomas Pock.
\newblock Search and rescue with airborne optical sectioning.
\newblock {\em IEEE Robotics and Automation Letters}, 5(2):2004--2011, 2020.

\bibitem{schedl2021airborne}
David~C. Schedl, Indrajit Kurmi, and Oliver Bimber.
\newblock Airborne optical sectioning for revealing persons in forested environments using synthetic aperture imaging.
\newblock {\em ISPRS Journal of Photogrammetry and Remote Sensing}, 175:332--346, 2021.

\bibitem{schedl2021autonomous}
David~C Schedl, Indrajit Kurmi, and Oliver Bimber.
\newblock An autonomous drone for search and rescue in forests using airborne optical sectioning.
\newblock {\em Science Robotics}, 6(55):eabg1188, 2021.

\bibitem{schoenberger2016mvs}
Johannes~Lutz Sch\"{o}nberger, Enliang Zheng, Marc Pollefeys, and Jan-Michael Frahm.
\newblock Pixelwise view selection for unstructured multi-view stereo.
\newblock In {\em European Conference on Computer Vision (ECCV)}, 2016.

\bibitem{schonberger2018robust}
Johannes~L. Schönberger.
\newblock {\em Robust Methods for Accurate and Efficient 3D Modeling from Unstructured Imagery}.
\newblock PhD thesis, ETH Zürich, 2018.

\bibitem{seits2022role}
Francis Seits, Indrajit Kurmi, Rakesh John Amala~Arokia Nathan, Rudolf Ortner, and Oliver Bimber.
\newblock On the role of field of view for occlusion removal with airborne optical sectioning.
\newblock {\em arXiv preprint arXiv:2204.13371}, 2022.

\bibitem{lichtfeld2025}
LichtFeld Studio.
\newblock A high-performance c++ and cuda implementation of 3d gaussian splatting, 2025.

\bibitem{thiel2020uas}
Christian Thiel, Marlin~M Mueller, Lea Epple, Christian Thau, S{\"o}ren Hese, Michael Voltersen, and Andreas Henkel.
\newblock Uas imagery-based mapping of coarse wood debris in a natural deciduous forest in central germany (hainich national park).
\newblock {\em Remote Sensing}, 12(20):3293, 2020.

\bibitem{vanvalkenburgh2020lasers}
Parker VanValkenburgh, KC Cushman, Luis Jaime~Castillo Butters, Carol~Rojas Vega, Carson~B Roberts, Charles Kepler, and James Kellner.
\newblock Lasers without lost cities: using drone lidar to capture architectural complexity at kuelap, amazonas, peru.
\newblock {\em Journal of Field Archaeology}, 45(sup1):S75--S88, 2020.

\bibitem{venier2019modelling}
Lisa~A Venier, Tom Swystun, Marc~J Mazerolle, David~P Kreutzweiser, Kerrie~L Wainio-Keizer, Ken~A McIlwrick, Murray~E Woods, and Xianli Wang.
\newblock Modelling vegetation understory cover using lidar metrics.
\newblock {\em PLoS One}, 14(11):e0220096, 2019.

\bibitem{wang2003multiscale}
Zhou Wang, Eero~P Simoncelli, and Alan~C Bovik.
\newblock Multiscale structural similarity for image quality assessment.
\newblock In {\em The thrity-seventh asilomar conference on signals, systems \& computers, 2003}, volume~2, pages 1398--1402. Ieee, 2003.

\bibitem{nasa2000ndvi}
John Weier and David Herring.
\newblock Measuring vegetation.
\newblock \url{https://earthobservatory.nasa.gov/features/MeasuringVegetation/measuring_vegetation_2.php}, 2000.
\newblock Accessed: 2025-07-01.

\bibitem{wu20253dgut}
Qi Wu, Janick~Martinez Esturo, Ashkan Mirzaei, Nicolas Moenne-Loccoz, and Zan Gojcic.
\newblock 3dgut: Enabling distorted cameras and secondary rays in gaussian splatting.
\newblock In {\em Proceedings of the Computer Vision and Pattern Recognition Conference}, pages 26036--26046, 2025.

\bibitem{Xue2015ObstructionFree}
Tianfan Xue, Michael Rubinstein, Ce Liu, and William~T. Freeman.
\newblock A computational approach for obstruction-free photography.
\newblock {\em ACM Transactions on Graphics (Proc. SIGGRAPH)}, 34(4), 2015.

\bibitem{yin2023or}
Youtan Yin, Zhoujie Fu, Fan Yang, and Guosheng Lin.
\newblock Or-nerf: Object removing from 3d scenes guided by multiview segmentation with neural radiance fields.
\newblock {\em arXiv preprint arXiv:2305.10503}, 2023.

\bibitem{zhu2023occlusion}
Chengxuan Zhu, Renjie Wan, Yunkai Tang, and Boxin Shi.
\newblock Occlusion-free scene recovery via neural radiance fields.
\newblock In {\em Proceedings of the IEEE/CVF Conference on Computer Vision and Pattern Recognition}, pages 20722--20731, 2023.

\end{thebibliography}

\normalsize
\end{document}